%% file: main.tex
\documentclass[runningheads]{llncs}
\usepackage{eccv}

\usepackage{eccvabbrv}

\usepackage{graphicx}
\usepackage{booktabs}
\usepackage{multirow}
\usepackage{float}
\usepackage{wrapfig}
\usepackage{dsfont}
\usepackage{algorithm}
\usepackage{algorithmic}
\usepackage{tcolorbox}

\usepackage[accsupp]{axessibility} 
\usepackage{hyperref}
\usepackage{orcidlink}

\begin{document}

\title{HO-Flow: Generalizable Hand-Object Interaction Generation with Latent Flow Matching} 

\titlerunning{HO-Flow}

\author{Zerui Chen$^{1}$ \and
Rolandos Alexandros Potamias$^{1}$ \and
Shizhe Chen$^{2}$\and \\ Jiankang Deng$^{1}$ \and Cordelia Schmid$^{2}$ \and Stefanos Zafeiriou$^{1}$\\[-0.6em]}

\authorrunning{Chen et al.}

\institute{$^{1}$Imperial College London \quad $^{2}$Inria, CNRS, ENS-PSL \\
\tt\small \url{https://zerchen.github.io/projects/hoflow.html}}

\maketitle
\begin{figure}[h]
\vspace{-0.7cm}
    \centering
    \includegraphics[width=0.91\textwidth]{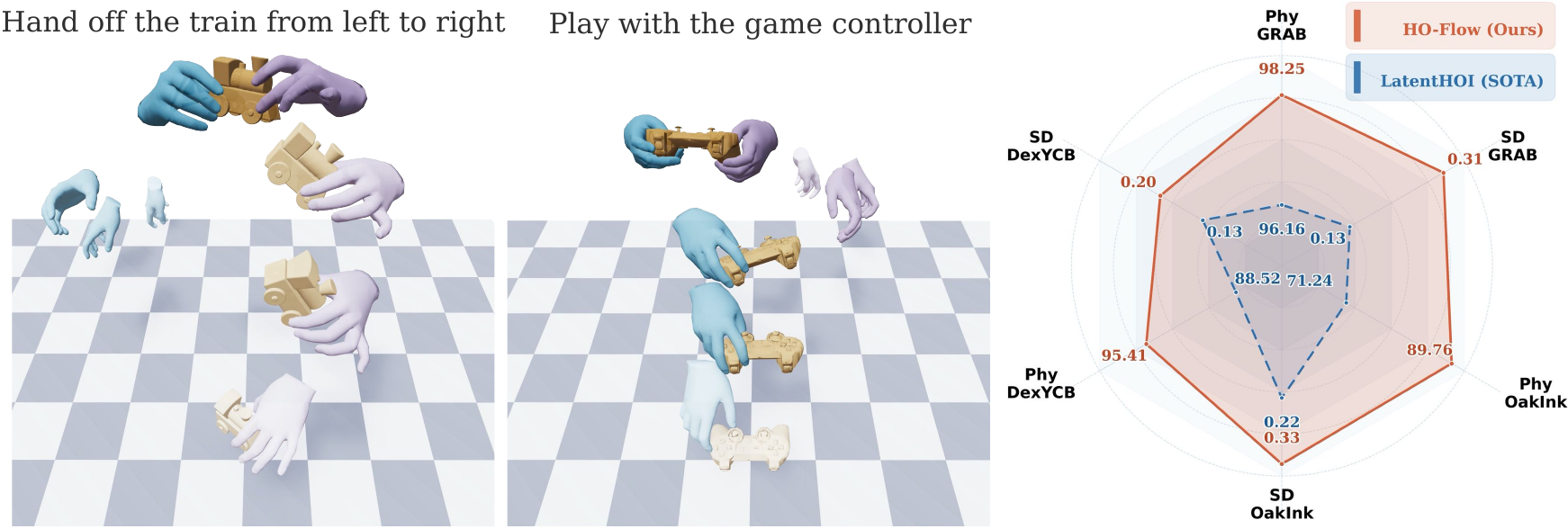}
    \vspace{-0.17cm}
    \caption{
    Our approach can synthesize realistic hand–object interactions for diverse tasks (left), where darker colors represent later time steps. It outperforms the state-of-the-art LatentHOI~\cite{Muchen_LatentHOI} across three benchmarks (right), namely  GRAB~\cite{GRAB:2020}, DexYCB~\cite{chao2021dexycb} and OakInk~\cite{yang2022oakink}. SD and Phy refer to sample diversity and physical plausibility.}
    \label{fig:teaser}
\end{figure}
\vspace{-1.1cm}

\begin{abstract}
Generating realistic 3D hand-object interactions (HOI) is a fundamental challenge in computer vision and robotics, requiring both temporal coherence and high-fidelity physical plausibility. 
Existing methods remain limited in their ability to learn expressive motion representations for generation and perform temporal reasoning.
In this paper, we present HO-Flow, a framework for synthesizing realistic hand-object motion sequences from texts and canoncial 3D objects.
HO-Flow first employs an interaction-aware variational autoencoder to encode sequences of hand and object motions into a unified latent manifold by incorporating hand and object kinematics, enabling the representation to capture rich interaction dynamics. 
It then leverages a masked flow matching model that combines auto-regressive temporal reasoning with continuous latent generation, improving temporal coherence.
To further enhance generalization, HO-Flow predicts object motions relative to the initial frame, enabling effective pre-training on large-scale synthetic data. Experiments on the GRAB, OakInk, and DexYCB benchmarks demonstrate that HO-Flow achieves state-of-the-art performance in both physical plausibility and motion diversity for interaction motion synthesis.

\vspace{-0.2cm}
\keywords{Hand-Object Interaction \and Latent Model \and Flow Matching}
\end{abstract}

\input{sections/introduction}
\input{sections/related_work}
\input{sections/method}
\input{sections/experiment}
\input{sections/conclusion}

{
    \small
    \bibliographystyle{splncs04}
    \bibliography{main}
}

\clearpage
\appendix
\section*{Appendix}
\input{appendix}
\end{document}

%% file: sections/introduction.tex
\section{Introduction}
\vspace{-0.1cm}
Generating realistic 3D hand-object interactions (HOI) is a fundamental challenge with wide-ranging applications in character animation, virtual reality, and robotic manipulation~\cite{dasari2023pgdm,luo2024omnigrasp}. 
While previous research has achieved impressive results in synthesizing static grasping poses\cite{jiang2021hand,li2024semgrasp,zhang2024nl2contact,liu2021synthesizing,miller2004graspit,zhu2025should}, generating dynamic, temporal HOI sequences from text~\cite{christen2024diffh2o} remains significantly more challenging and relatively under-explored.
Extending generative frameworks to the temporal 4D domain introduces substantial difficulties, including maintaining physical plausibility, preserving temporal coherence over long time horizons, and achieving generalization across diverse actions and objects.

A central challenge lies in preventing physical inconsistencies, particularly interpenetration artifacts where the hand geometry passes through the manipulated object during motion. 
Addressing this issue requires a \emph{highly expressive motion representation} capable of modeling the coupled dynamics between the hand and the object. 
Early approaches~\cite{christen2024diffh2o} adopt diffusion models~\cite{ho2020ddpm} to directly generate sequences of raw hand and object poses. 
However, the high degree of freedom of articulated hand models result in considerable computational cost, and treating poses as unconstrained independent variables often fails to faithfully model the underlying interaction distribution, producing artifacts such as floating hands or implausible contacts.
More recent methods, such as LatentHOI~\cite{Muchen_LatentHOI}, alleviate this issue by embedding hand-object poses into a \emph{compact latent space}. Nevertheless, existing latent representations are typically learned frame-independently, which weakens temporal coherence and does not encode fine-grained, contact-aware cues that are necessary for high-fidelity synthesis.

Beyond representation, the generative architecture itself plays a crucial role. 
\emph{Auto-regressive (AR) Transformers}~\cite{huang2025hoigpt} are effective at modeling long-range temporal dependencies, but they typically require quantizing continuous motions into discrete tokens~\cite{van2017neural}. Such discretization can suppress contact-rich features that are essential for realistic manipulation. 
In contrast, \emph{diffusion and flow-based models}~\cite{christen2024diffh2o, Muchen_LatentHOI} operate directly in continuous space, thereby avoiding quantization artifacts. 
However, these models usually perform holistic denoising over entire sequences, which incurs substantial computational cost. As a result, they often rely on U-Net architectures rather than more expressive Transformer-based models, and tend to produce globally inconsistent motions.

In this work, we present HO-Flow to synthesize realistic hand-object motion sequences based on text prompts.
First, we learn an expressive motion representation using a novel \emph{interaction-aware variational autoencoder (Inter-VAE)}. Unlike frame-wise motion encoders, Inter-VAE embeds short-horizon temporal sequences of hand-object interactions in latent spaces. To enrich fine-grained interaction cues, we transform object point clouds into hand-centric coordinates using the hand kinematic chain, and capture both global motion trajectories and contact-rich hand-object coordination in the latent.
Building upon this representation, we introduce a masked flow matching model to generate motion latents. This design attends to the full temporal context via the auto-regressive masked transformer while operating in continuous space via a lightweight flow matching head, significantly reducing jitter and outlier frames while promoting long-range coherence and smoothness.
To further improve generalization, HO-Flow predicts object motions relative to the initial frame, avoiding dataset-specific coordinate conventions, which enables effective pre-training on large-scale synthetic data~\cite{zhang2024graspxl} and yields superior transfer to unseen objects. 
We evaluate HO-Flow on three challenging benchmarks, including GRAB~\cite{GRAB:2020}, OakInk~\cite{yang2022oakink}, and DexYCB~\cite{chao2021dexycb}.
Experimental results demonstrate that HO-Flow advances the state of the art in terms of both physical plausibility and motion diversity.

\noindent Our contributions are summarized as follows:

\noindent$\bullet$ We propose HO-Flow for text-driven hand–object interaction motion generation, combining an expressive motion representation with an efficient temporal generative modeling to synthesize realistic and coherent HOI sequences.

\noindent$\bullet$ We introduce Inter-VAE to enhance motion representation learning, which encodes temporal interaction sequences into a unified latent space, capturing both global motion trajectories and fine-grained contact-rich coordination.

\noindent$\bullet$ HO-Flow achieves state-of-the-art performance on three challenging benchmarks for hand–object motion generation, benefiting from its expressive motion representation, auto-regressive temporal reasoning, and large-scale pre-training.

%% file: sections/related_work.tex
\section{Related Work}
\noindent \textbf{Hand grasp generation.}
Synthesizing realistic hand grasps for 3D objects is a long-standing problem in computer vision and robotics~\cite{bohg2013data}. Early approaches rely on optimization to find physically plausible grasps given known object geometry~\cite{miller2004graspit,liu2021synthesizing,du2022multi}. GraspIt!~\cite{miller2004graspit}, for instance, formulates grasp synthesis as an optimization problem over hand configurations to maximize a physically grounded quality metric. Subsequent works incorporate force-closure constraints~\cite{liu2021synthesizing,zhong2025dexgrasp} and biomechanical contact priors~\cite{yang2022oakink,yang2021cpf} to further improve realism. With the advent of deep learning, data-driven methods train conditional variational autoencoders~\cite{GRAB:2020,karunratanakul2020grasping,wu2025cedex,lu2024ugg} or transformers~\cite{xu2024dexterous,wei2024grasp} to generate hand grasps conditioned on object shape, yielding diverse and plausible configurations. More recent works introduce richer conditioning signals: NL2Contact~\cite{zhang2024nl2contact} and SemGrasp~\cite{li2024semgrasp} leverage natural language and semantic cues to guide synthesis toward task-relevant grasps, while affordance-based methods~\cite{ye2023affordance,jiang2021hand,wei2025afforddexgrasp} explicitly model the relationship between object geometry and plausible hand placements. Despite their impressive results, all these methods predict only static hand grasping poses for objects. In this work, we address the more challenging problem of generating dynamic hand-object interaction sequences.

\noindent \textbf{Hand and object motion generation.} Generating hand-object interaction motions from task descriptions has attracted growing attention, with approaches broadly divided into optimization-based and learning-based methods. Unlike human motion generation~\cite{zhang2023generating,petrovich22temos,athanasiou23sinc}, this task emphasizes physical naturalness for interactions. Optimization-based methods leveraging reinforcement learning~\cite{sutton2018reinforcement,schulman2017proximal,xu2023unidexgrasp,wan2023unidexgrasp++} or physics-based constraints~\cite{zhang2025bimart,ye2025contact2motion} can yield physically plausible results, but require hours of per-sequence optimization and tend to produce monotonous motions, limiting their scalability. Among learning-based approaches, HOI-GPT~\cite{huang2025hoigpt} auto-regressively generates discrete motion tokens via VQ-VAE~\cite{van2017neural}, through quantizing continuous poses into a finite codebook inherently limits motion diversity and fine-grained interaction fidelity. Diffusion models~\cite{ho2020ddpm} avoid this bottleneck by operating in continuous spaces. DiffH2O~\cite{christen2024diffh2o} directly denoises explicit hand and object pose sequences, but jointly modeling 3D positions and rotations that lie on incompatible manifolds renders the problem ill-posed and limits motion quality. LatentHOI~\cite{Muchen_LatentHOI} mitigates this by encoding poses into a latent space before applying diffusion, yet its frame-by-frame encoding captures only local grasping poses while neglecting global hand trajectories and object motions, resulting in temporally inconsistent and jittery outputs. In contrast, HO-Flow holistically encodes full hand-object interaction sequences into a unified structured latent space and auto-regressively generates temporally coherent and fine-grained interaction motions.

\noindent \textbf{Diffusion and flow matching models.}
Diffusion models have emerged as a powerful paradigm for high-fidelity generative modeling across images, videos, and human motions.
They define a forward process that gradually perturbs data into noise, and learn a reverse denoising process to sample new instances~\cite{ho2020ddpm,song2021scorebased,sohl2015deep}.
This framework has driven substantial progress in image synthesis, especially when combined with improved noise schedules and parameterizations~\cite{nichol2021improved,karras2022edm}, and scalable latent-space formulations for efficient high-resolution generation~\cite{rombach2022ldm}.
Extending diffusion to videos introduces additional challenges in modeling spatio-temporal coherence, and recent works tackle this via temporally-aware architectures and conditioning strategies~\cite{ho2022video,blattmann2023videoldm}.
Diffusion-based motion generators similarly benefit from operating in continuous spaces and have shown strong performance for text-conditioned human motion synthesis, thanks to their ability to model multi-modal motion distributions and long-range temporal dependencies~\cite{tevet2022human,zhang2022motiondiffuse}.
In parallel, flow-based continuous-time generative models provide an alternative that learns a velocity field to transport noise into data via ODE integration.
Flow matching~\cite{lipman2023flowmatching} and rectified flow~\cite{liu2023rectified} unify and simplify training objectives for such continuous-time generators, enabling faster sampling with fewer integration steps while retaining high sample quality.
Motivated by these advances, we introduce HO-Flow, which uses interaction-aware latents and masked flow matching for efficient and coherent hand-object motion generation.

%% file: sections/method.tex
\section{Method}
\label{sec:method}

In this section, we introduce HO-Flow (Figure~\ref{fig:overview}), a novel framework that generates Hand–Object interaction motions from task descriptions using an auto-regressive Flow matching model. We first outline the overall architecture (Section~\ref{sec:method_overview}), then detail two core components: (i) an interaction-aware VAE (Section~\ref{sec:method_vae}) that encodes explicit hand–object poses into compact motion tokens, and (ii) a context-aware auto-regressive flow matching model (Section~\ref{sec:method_ar}) that generates coherent interaction sequences conditioned on task specifications.

\subsection{Overview}
\label{sec:method_overview}
\begin{wrapfigure}[15]{r}{0.6\linewidth}
  \centering
  \includegraphics[width=1\linewidth]{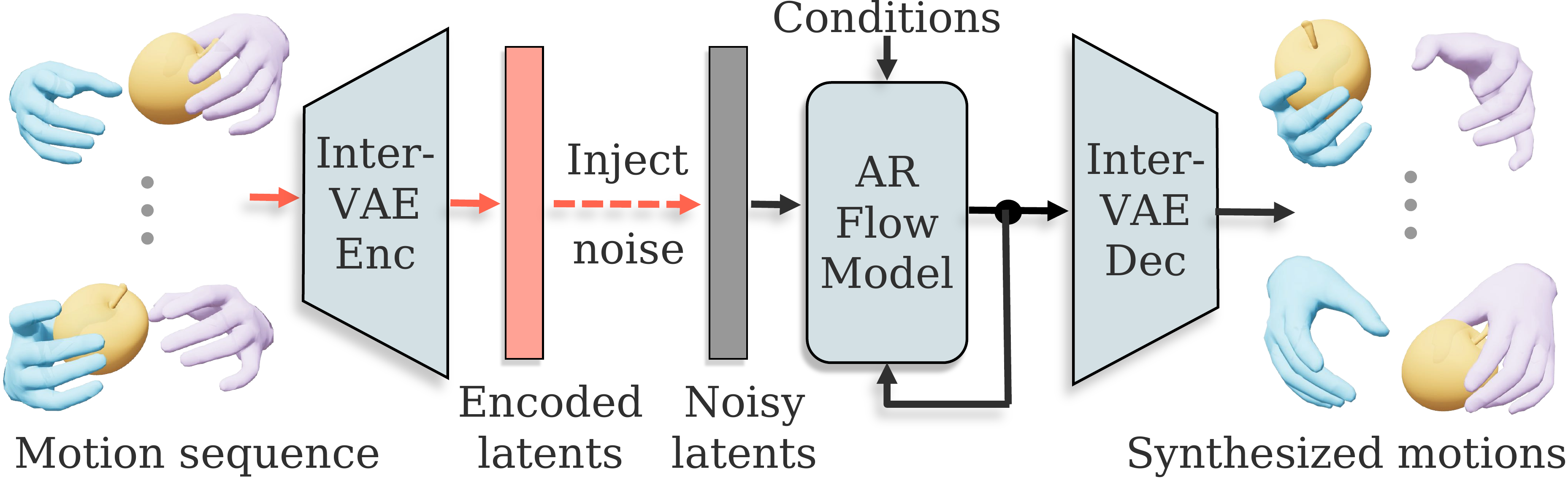}
  \caption{Overview of HO-Flow. Red and gray blocks indicate encoded and noisy motion latents, respectively. Red arrows are active only during training. The framework synthesizes realistic hand-object interactions using two components: an interaction-aware VAE for compact motion latents with fine-grained interaction features, and an auto-regressive flow-matching model that predicts successive latents for faithful, temporally coherent synthesis.}
  \label{fig:overview}
  \vspace{-0.5\baselineskip}
\end{wrapfigure}

Figure~\ref{fig:overview} shows HO-Flow, a novel approach for realistic hand-object interaction synthesis conditioned on the text description and the object point cloud.
Realistic hand-object interaction synthesis requires capturing contact-rich dynamics and long-range temporal dependencies beyond frame-wise pose representations. To this end, the proposed method introduces two coupled components: (1) an interaction-aware VAE that compresses hand-object dynamics into structured latents for holistic sequence-level reasoning, and (2) a flow matching model that auto-regressively generates these latents in continuous space for efficient, temporally coherent synthesis.

The interaction-aware VAE employs a symmetric encoder-decoder architecture. The encoder leverages an object point cloud alongside hand and object motion sequences to learn unified latent representations. Within the encoder, a spatial module first extracts per-frame interaction features, which are subsequently aggregated by temporal encoders to produce distinct latent codes for the hand and the object. To regularize the latent space and ensure reconstruction fidelity, dedicated hand and object decoders are trained to recover the original motion sequences from these learned spatial-temporary interaction features.

Following the training of the motion latents, we train an auto-regressive flow matching model. Conditioned on a sparse object point cloud and a natural language description, this model utilizes a masked transformer architecture and auto-regressively aggregates motion context to predict successive motion tokens via a flow matching objective. During inference, the frozen decoders from the interaction-aware VAE decode the generated latent codes back into the space of hand and object poses.

\subsection{Interaction-aware Variational Autoencoder}
\label{sec:method_vae}

\begin{wrapfigure}[12]{r}{0.6\linewidth}
  \centering
  \vspace{-2.7\baselineskip}
  \includegraphics[width=\linewidth]{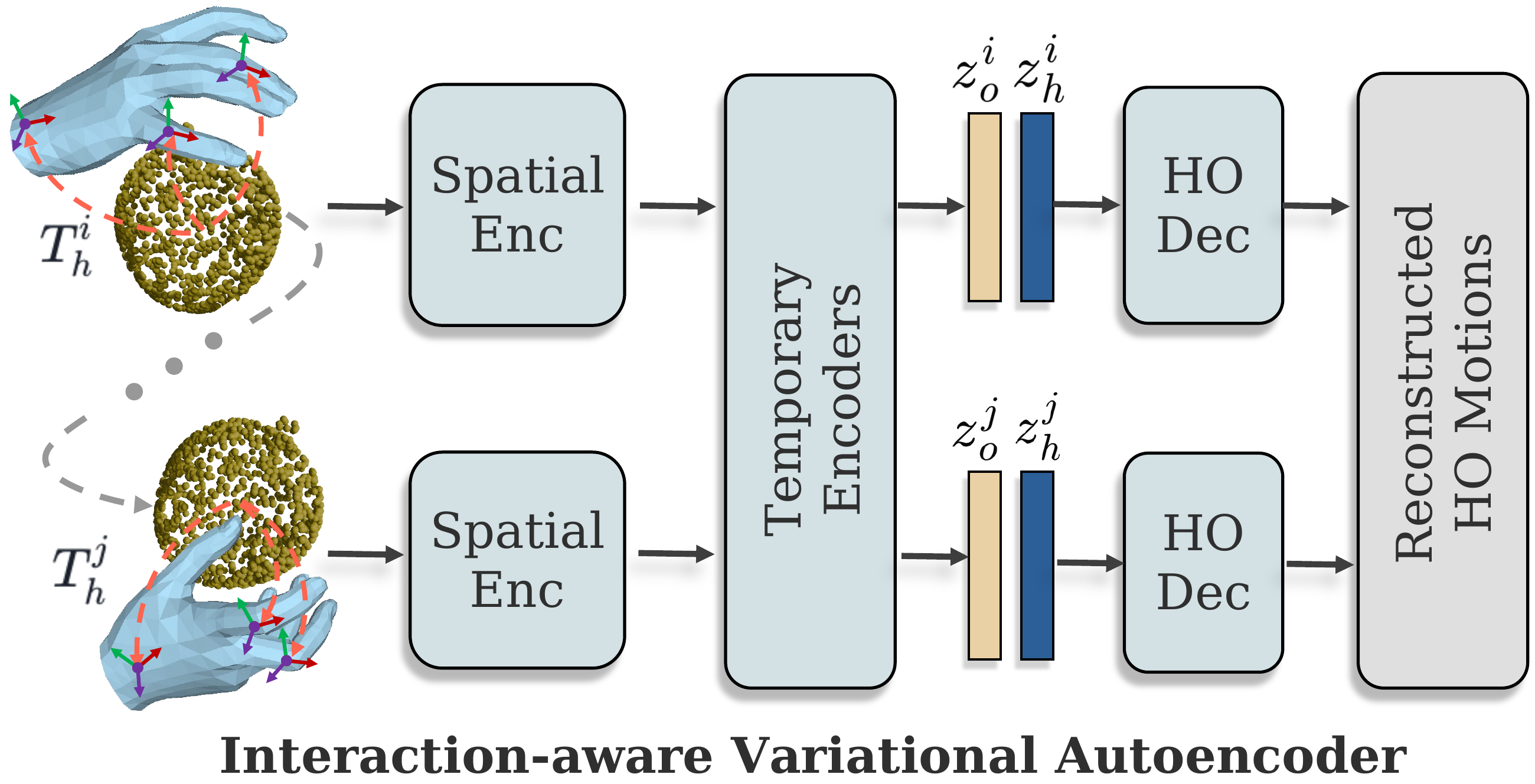}
  \caption{Interaction-aware VAE captures fine-grained interaction features for latent representations ($\mathbf{z}_{o}$ and $\mathbf{z}_{h}$). $\mathbf{T}_{h}$ is the transformation of hand bones.}
  \label{fig:method_vae}
  \vspace{-0.8\baselineskip}
\end{wrapfigure}

As illustrated in Figure~\ref{fig:method_vae}, we propose an interaction-aware variational autoencoder (VAE) to compress hand-object interaction sequences into compact, structured latent representations. The VAE takes the object point cloud $\mathbf{p}_{o} \in \mathbb{R}^{1024\times3}$, MANO~\cite{MANO:SIGGRAPHASIA:2017} hand poses $\boldsymbol{\theta}_{h} \in \mathbb{R}^{\rm N\times16\times3}$, and object poses $\mathbf{T}_{o} \in \mathbb{R}^{\rm N\times4\times4}$ as inputs, where $\rm N$ denotes the number of motion frames. To mitigate the scarcity of bi-manual manipulation data, we follow~\cite{Muchen_LatentHOI} by focusing on the object manipulation with the right hand and mirroring left-hand data when available.

\noindent \textbf{Kinematic-aware object transformation.} To capture fine-grained interaction features, we project the object point clouds into the local coordinate systems of various hand joints. We first transform the object point clouds into the world coordinate system, $\mathbf{p}_{o}^{w} \in \mathbb{R}^{\rm N\times1024\times3}$, using the pose sequence $\mathbf{T}_{o}$. We then derive the global transformation $\mathbf{T}_{h, t}^{i} \in \mathbb{R}^{4\times 4}$ for the $i$-th hand joint using its pose $\boldsymbol{\theta}_{h,t}^{i}$ and the MANO kinematic tree:
\begin{equation}
\mathbf{T}_{h,t}^{i} = \prod_{j \in A(i)}\left[\begin{array}{c|c}
\exp \left (\boldsymbol{\theta}_{h,t}^{j}\right) & \boldsymbol{\phi}_{h}^{j} \\
\hline \mathbf{0} & 1
\end{array}\right],
\label{mano_kine}
\end{equation}
where $A(i)$ is the ordered set of ancestors of the $i$-th joint and $\exp(\cdot)$ is the Rodrigues formula converting axis-angle $\boldsymbol{\theta}_{h, t}^{j}$ to a rotation matrix. The translation $\boldsymbol{\phi}_{h}^{j}$ is the joint's offset relative to its parent as defined in the MANO template.

\noindent \textbf{Multi-view interaction features.} By traversing the kinematic chain, we obtain global transformations for all joints. We then transform the world-frame object point clouds $\mathbf{p}_{o}^{w}$ into each joint's local frame via $(\mathbf{T}_{h, t}^{i})^{-1}$ and concatenate them to form $\mathbf{p}_{o}^{h} \in \mathbb{R}^{N\times1024\times48}$, encoding rich hand-relative geometric context:
\begin{equation}
\mathbf{p}_{o}^{h} = \bigoplus_{i=0}^{15} \widetilde{\rm H}((\mathbf{T}_{h, t}^{i})^{-1} \cdot {\rm H}(\mathbf{p}_{o}^{w})),
\label{inv_trans}
\end{equation}
where $\bigoplus$ denotes concatenation, $\rm H(\cdot)$ denotes conversion to homogeneous coordinates, and $\widetilde{\rm H}(\cdot)$ denotes inverse projection back to Euclidean space. 

\noindent \textbf{Spatial-Temporal encoding.} The concatenated point clouds $\{\mathbf{p}_{o}, \mathbf{p}_{o}^{w}, \mathbf{p}_{o}^{h}\}$ are processed by a spatial encoder based on the PointNet++ architecture~\cite{qi2017pointnet++}, utilizing three set abstraction layers to extract features $\mathbf{f}_{s} \in \mathbb{R}^{N\times768}$. These features are subsequently fused with $\boldsymbol{\theta}_{h}$ and $\mathbf{T}_{o}$ via two MLP layers to yield hand and object spatial features $\mathbf{f}_{h}, \mathbf{f}_{o} \in \mathbb{R}^{256}$, respectively. Finally, temporal encoders, comprised of 1D convolutional layers with a total stride of 4, aggregate these features across the sequence to produce compact latent codes $\mathbf{z}_{h}, \mathbf{z}_{o} \in \mathbb{R}^{\frac{\rm N}{4}\times32}$.

\noindent \textbf{Motion Reconstruction.} To optimize the latent space, dedicated hand and object decoders upsample the latent features to the original sequence length and utilize convolutional layers to reconstruct the hand and object motion trajectories. We supervise the VAE using reconstruction losses over both hand and object motions, combined with KL regularization on the latent posteriors. The overall objective is formulated as:
\begin{equation}
\begin{split}
\mathcal{L}_{\text{VAE}} = \mathcal{L}_{\text{pose}} + \mathcal{L}_{\text{trans}} + \mathcal{L}_{\text{mesh}}
+ \mathcal{L}_{\text{obj-rot}} + \mathcal{L}_{\text{obj-trans}}
+ \beta\left(\mathcal{L}_{\text{KL}}^{h} + \mathcal{L}_{\text{KL}}^{o}\right),
\end{split}
\label{eq:vae_loss}
\end{equation}
where we utilize $\ell_1$ losses for all reconstruction terms and set $\beta$ to 1e-4. $\mathcal{L}_{\text{pose}}$ penalizes errors in the 6D rotation representation~\cite{zhou2019continuity} for all MANO joints. $\mathcal{L}_{\text{trans}}$ supervises the hand translation predicted relative to the object translation to encourage stable interaction geometry. $\mathcal{L}_{\text{mesh}}$ enforces vertex-level consistency by reconstructing MANO meshes. For the object, $\mathcal{L}_{\text{obj-rot}}$ and $\mathcal{L}_{\text{obj-trans}}$ supervise the object's 6D rotation and translation. The KL terms regularize the temporally-strided latent sequences:
\begin{equation}
\mathcal{L}_{\text{KL}}^{(\cdot)} = \mathrm{KL}\!\left(q(\mathbf{z}_{(\cdot)} \mid \mathbf{p}_o, \boldsymbol{\theta}_h, \mathbf{T}_o)\,\|\,\mathcal{N}(\mathbf{0}, \mathbf{I})\right), \quad (\cdot)\in\{h,o\}.
\label{eq:vae_kl}
\end{equation}

\subsection{Auto-regressive Flow Matching}
\label{sec:method_ar}
\begin{figure}[t]
  \centering  
  \includegraphics[width=1.0\linewidth]{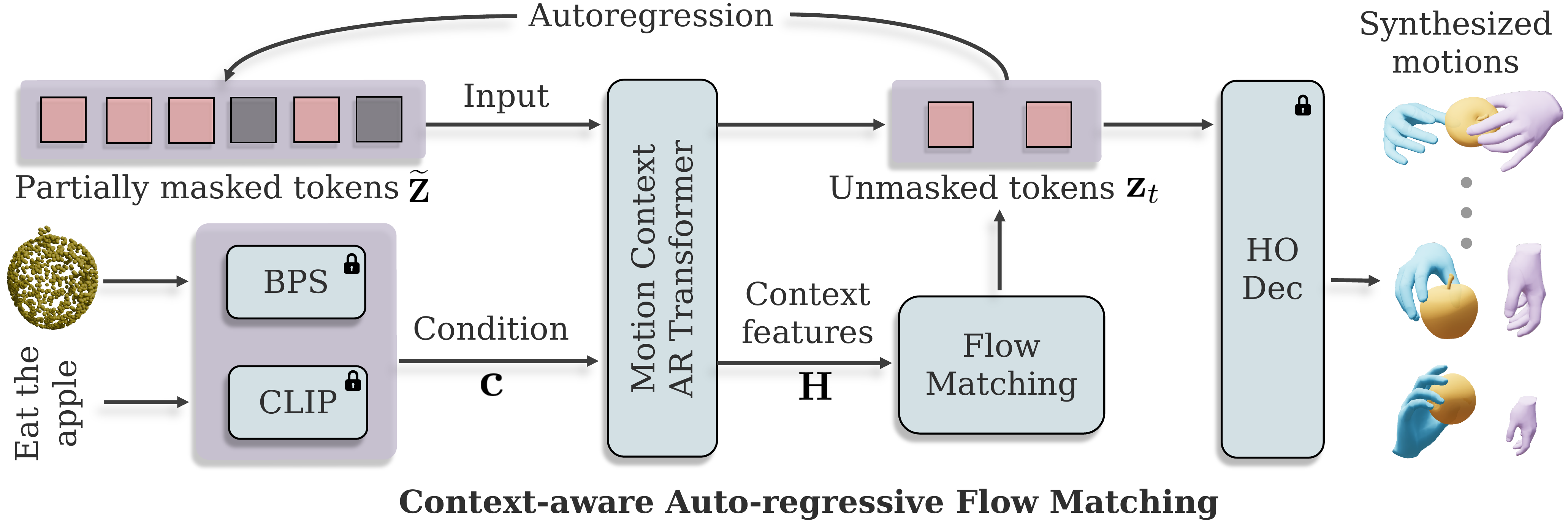}
  \vspace{-0.3cm}
  \caption{Illustration of our flow matching model. Gray and pink blocks indicate masked and unmasked tokens, respectively. Conditioned on object point clouds and task descriptions, the model aggregates motion context to autoregressively predict successive motions, thereby reducing uncertainty and enabling high-fidelity interaction synthesis.}
  \vspace{-0.4cm}
  \label{fig:method_fm}
\end{figure}

As illustrated in Figure~\ref{fig:method_fm}, we employ auto-regressive flow matching to generate interaction motion latents conditioned on the object geometry $\mathbf{p}_o$ and natural-language task description $s$. Built on the structured latent space $\mathbf{Z}$ learned by our interaction-aware VAE, it models $p(\mathbf{Z} \mid \mathbf{p}_o, s)$ to produce temporally coherent and realistic hand-object interactions. 

Given an input sequence of $\rm N$ frames, the temporal stride of the VAE yields a latent sequence of length $L = {\rm N}/4$. For each temporal step $t$, we concatenate the object and hand latents for both the right hand and its mirrored left counterpart to form a unified motion token:
\begin{equation}
\begin{split}
\mathbf{z}_t &= \left[\mathbf{z}^{r}_{o,t}\;\Vert\;\mathbf{z}^{r}_{h,t}\;\Vert\;\mathbf{z}^{l}_{o,t}\;\Vert\;\mathbf{z}^{l}_{h,t}\right]\in\mathbb{R}^{128}, \\
\mathbf{Z} &= [\mathbf{z}_1,\dots,\mathbf{z}_L]\in\mathbb{R}^{\frac{\rm N}{4}\times 128},
\end{split}
\label{eq:token_def}
\end{equation}
where $\mathbf{z}^{(\cdot)}_{o,t}$, $\mathbf{z}^{(\cdot)}_{h,t} \in \mathbb{R}^{32}$ denote the object and hand latent codes, respectively.

\noindent \textbf{Condition encoding.} We encode the task description $s$ using a frozen CLIP~\cite{radford2021learning} text encoder to obtain a semantic embedding $\mathbf{e}_{\text{text}} \in \mathbb{R}^{512}$. To represent the object geometry $\mathbf{p}_o$ as a fixed-length descriptor, we follow previous practice~\cite{Muchen_LatentHOI,christen2024diffh2o} to compute a Basis Point Set (BPS)~\cite{prokudin2019efficient} encoding, resulting in distance features $\mathbf{e}_{\text{bps}} \in \mathbb{R}^{4096}$. These features are projected into a shared conditioning space via linear layers and summed to form the final condition vector:
\begin{equation}
\mathbf{c} = \mathbf{W}_{\text{text}}\mathbf{e}_{\text{text}} + \mathbf{W}_{\text{bps}}\mathbf{e}_{\text{bps}} \in \mathbb{R}^{d},
\label{eq:cond_embed}
\end{equation}
where we set $d$ to 1024 in the implementation.

\noindent \textbf{Context-aware transformer.} To capture long-range temporal dependencies while facilitating progressive generation, we adopt a masked auto-regressive transformer (MAR) architecture over the latent tokens. The sequence of motion tokens $\mathbf{Z}$ is embedded, augmented with sinusoidal positional encodings, and processed by a stack of transformer blocks. Each block utilizes adaptive LayerNorm (AdaLN)~\cite{peebles2023scalable} modulation conditioned on $\mathbf{c}$ to yield contextual features $\mathbf{H} = \mathrm{MAR}(\widetilde{\mathbf{Z}}, \mathbf{c}) \in \mathbb{R}^{\frac{\rm N}{4} \times d}$, where $\widetilde{\mathbf{Z}}$ represents the partially masked input sequence during training or inference.

\noindent \textbf{Flow matching in latent space.} For each masked latent position $t \in \mathcal{M}$, we learn a conditional flow matching model to map a prior noise distribution to the target token $\mathbf{z}_t$. We define a probability path as a straight-line linear interpolation between noise $\mathbf{x}_0 \sim \mathcal{N}(\mathbf{0}, \mathbf{I})$ and the ground-truth latent $\mathbf{x}_1 = \mathbf{z}_t$:
\begin{equation}
\mathbf{x}_\tau = \tau \mathbf{x}_1 + (1-\tau)\mathbf{x}_0, \quad \tau \in [0,1].
\label{eq:fm_path}
\end{equation}
The corresponding conditional vector field is the time-derivative of this path, which serves as the ground-truth supervision target for our model:
\begin{equation}
\mathbf{u}_t(\mathbf{x}_\tau) = \frac{d\mathbf{x}_\tau}{d\tau} = \mathbf{x}_1 - \mathbf{x}_0.
\label{eq:velocity}
\end{equation}
We parameterize a velocity field $v_{\phi}(\mathbf{x}_\tau, \tau; \mathbf{h}_t)$ using a lightweight MLP conditioned on the contextual features $\mathbf{h}_t$. The model is trained to regress the constant velocity $\mathbf{u}_t$ by minimizing the flow matching objective. This formulation effectively guides noise along a deterministic trajectory toward the hand-object interaction manifold, enabling efficient inference via ODE solvers. 

In training, we pre-compute latent targets $\mathbf{Z}$ from the dataset with the frozen VAE. The model is trained to inpaint masked tokens using a masked modeling strategy. Specifically, we sample a subset of token positions $\mathcal{M}$ according to a cosine mask-ratio schedule. These positions are corrupted using a BERT-style strategy: 80\% are replaced by a learnable \texttt{[MASK]} token, 10\% by random Gaussian noise, and 10\% are left unchanged. For each masked position $t \in \mathcal{M}$, we minimize the flow matching objective:
\begin{equation}
\mathcal{L}_{\text{AR}} = \mathbb{E}_{\mathcal{M}, \tau, \mathbf{x}_0}\left[\frac{1}{|\mathcal{M}|}\sum_{t\in\mathcal{M}} \left\| v_{\phi}(\mathbf{x}_\tau, \tau; \mathbf{h}_t) - (\mathbf{z}_t - \mathbf{x}_0) \right\|_2^2 \right],
\label{eq:ar_loss}
\end{equation}
where $\mathbf{h}_t$ is the contextual feature extracted by the MAR transformer from the corrupted sequence. We apply classifier-free guidance~\cite{ho2022classifier} by randomly dropping the text condition during training ($20\%$ probability) and maintain an exponential moving average (EMA) of model parameters for stable inference.

%% file: sections/experiment.tex
\section{Experimental Evaluation}
\label{sec:exp}

We conduct comprehensive experiments on three hand-object motion benchmarks to validate the effectiveness of HO-Flow in synthesizing both bi-manual and single-hand manipulation motions across diverse objects.

\subsection{Datasets}
\label{subsec:data}

\noindent \textbf{GRAB}~\cite{GRAB:2020}: The dataset contains bimanual manipulation tasks. Following the object-based split in~\cite{Muchen_LatentHOI}, we reserve 4 objects for testing and 47 for training. 
To focus on the manipulation phase, training sequences begin at the first contact frame, and sequences are padded or truncated to a maximum of 160 frames. 
The unseen test split consists of 17 (text prompt, object) pairs.

\noindent \textbf{OakInk}~\cite{yang2022oakink}: To evaluate out-of-distribution generalization, we use a challenging novel-object split~\cite{Muchen_LatentHOI} from this dataset, selecting 100 objects across 20 categories. These unseen shapes are paired with intents from GRAB, resulting in 212 evaluation pairs. As no motion data is used for training on this split, we use the model trained on GRAB for this evaluation setup.

\noindent \textbf{DexYCB}~\cite{chao2021dexycb}: For single-hand grasping, we employ an object-based split by reserving 4 unseen objects out of 20 for testing. Sequences start from the first frame and are padded to a maximum of 96 frames.

\noindent \textbf{GraspXL}~\cite{zhang2024graspxl}: It provides over five million synthetic right-hand manipulation trajectories across half a million diverse objects, generated via RL in a physics-based simulator. 
While physically plausible, it is restricted to relocation tasks and lacks motion diversity. 
We therefore leverage this large-scale data for pre-training to learn interaction priors before fine-tuning on more complex motions.

\subsection{Evaluation metrics}
\label{subsec:metric}
We evaluate our method in terms of latent representation quality and the physical plausibility and diversity of generated interaction motions, following prior works~\cite{Muchen_LatentHOI,karunratanakul2020grasping,christen2024diffh2o,chen2022alignsdf}. Please refer to the appendix for details of different metrics.

To evaluate the quality of the learned latent representation, we report mean joint error ($\rm E_{j}$) and mean vertex error ($\rm E_{v}$) in mm for the hand, and mean translation error ($\rm E_{o}$) in mm and chamfer distance ($\rm CD_{o}$) in cm$^2$ for the object, where $\rm CD_{o}$ also reflects orientation error and is robust to object symmetries.

To evaluate the physical plausibility and diversity of generated interaction motions, we adopt the following evaluation metrics for motion generation.

\noindent \textbf{Contact ratio (${\rm \bf{CR}}$)}. We compute the percentage of hand vertices in contact with the object surface and report the average ratio (\%) over the sequence.

\noindent \textbf{Interpenetration (${\rm \bf{IV/ID}}$)}. To quantify collisions, we report the hand-object intersection volume (${\rm IV}$, cm$^3$) and the maximum penetration depth (${\rm ID}$, cm), both averaged over frames with non-zero interpenetration.

\noindent \textbf{Physics score (${\rm \bf{IVU/Phy}}$)}. We report the interpenetration volume normalized by the estimated contact area (${\rm IVU}$) and the percentage of frames, Phy (\%), in which the object moves while maintaining hand-object contact.

\noindent \textbf{Sample diversity (${\rm \bf{SD}}$)}. We compute sample diversity as the average pairwise $\ell_2$ distance among multiple generations conditioned on the same input.

\subsection{Implementation details}
\label{subsec:details}
\noindent \textbf{Model architecture.}
We implement the two-stage HO-Flow framework described in Section~\ref{sec:method}. In the interaction-aware VAE, temporal encoders and decoders comprise two stride-2 downsampling and upsampling stages with a hidden dimension of 512. For the context-aware autoregressive model, we encode text features using a frozen CLIP ViT-B/32 and employ a single-layer transformer with 16 attention heads to efficiently encode the context features $\mathbf{h}_t$. Additional architectural details are provided in appendix.

\noindent \textbf{Training and inference Details.} 
We optimize all models using AdamW~\cite{loshchilov2017decoupled} with linear warmup and cosine learning rate decay. The interaction-aware VAE is first pre-trained on GraspXL~\cite{zhang2024graspxl} for 100k iterations with a batch size of 32 and a base learning rate of $2\times10^{-4}$ (decayed to $2\times10^{-5}$) across 4 NVIDIA H100 GPUs. Then, it is fine-tuned on the specific dataset under the same setting but with a base learning rate of  $1\times10^{-4}$. To enhance robustness, we apply random global 3D rotations to the whole manipulation sequence as data augmentation. Once converged, the VAE is frozen to pre-compute latent targets for the generative stage. The latent flow matching model is also first pre-trained on GraspXL and subsequently fine-tuned on specific dataset for 300k iterations with a batch size of 32 on a single NVIDIA H100 GPU, following the same warmup and decay schedules used for Inter-VAE. We employ an Exponential Moving Average (EMA) with a decay of 0.9999 for stability. During inference, we generate samples using 18 steps with a classifier-free guidance weight of 1.5.

\subsection{Ablation studies}
\label{subsec:ablation}
We conduct comprehensive ablation studies on the GRAB dataset to evaluate the effectiveness of each component of our proposed HO-Flow approach.

\input{tables/abl_vae}

\noindent \textbf{Interaction-aware variational autoencoder.} 
Table~\ref{tab:iv} presents an ablation study of the proposed components within our VAE model. 
The baseline (\textbf{R1}), which utilizes canonical object point clouds alongside hand and object poses, fails to capture the spatial interactions in world coordinates, leading to unsatisfactory performance. 
By incorporating 6D object poses and processing object point clouds directly in the world coordinate space (\textbf{R2}), we observe a significant reduction in object reconstruction errors ($\rm E_o$ and $\rm CD_o$). 
Transforming these point clouds into various hand-joint coordinate systems (\textbf{R3}) further enhances the extraction of hand-object interaction features, yielding consistent improvements across both hand and object metrics. 
In \textbf{R4}, we transition from a shared latent code to separate latent representations for the hand and object. This decoupling facilitates the extraction of more discriminative motion features, further boosting accuracy. 
Finally, pre-training Inter-VAE on synthetic trajectories from GraspXL~\cite{zhang2024graspxl} (\textbf{R5}) significantly enhances model generalization to unseen objects, resulting in the best performance across all metrics.

\input{tables/abl_fm}

\noindent \textbf{Auto-regressive latent flow matching.} We conduct ablation studies on several variants of the generative model architecture, with results summarized in Table~\ref{tab:fm}. The baseline \textbf{R1} directly generates hand and object motions in the original pose space. This design leads to limited physical fidelity, achieving the lowest physical plausibility score (84.89 in Phy) and severe right-hand interpenetration (10.35 in $\rm IV_r$). We then consider \textbf{R2}, which adopts the latent representation of LatentHOI~\cite{Muchen_LatentHOI}. Specifically, it encodes motion frames independently and uses a VAE to embed local hand poses into a latent space while keeping global hand and object motions in explicit pose space. Without temporal latent modeling, \textbf{R2} struggles to capture holistic hand-object interaction patterns across the sequence, resulting in poor temporal coherence and physical realism (\emph{e.g.}, 8.13 in $\rm IV_r$ and 88.76 in Phy), as well as reduced motion diversity (0.12 in SD). Compared with \textbf{R2}, \textbf{R3} introduces our proposed interaction-aware VAE (Section~\ref{sec:method_vae}) to jointly encode hand-object interaction features in both spatial and temporal domains. This expressive latent representation improves overall physical fidelity (Phy increases to $92.69$), reduces penetration and distance errors (\emph{e.g.}, $\rm IV_r$ drops to $6.99$, $\rm ID_r$ drops to $0.82$), and substantially improves diversity (SD increases to $0.26$). Finally, \textbf{R4} incorporates our context-aware transformer (Section~\ref{sec:method_ar}) with an auto-regressive formulation, where each token is predicted conditioned on the generated motion context rather than predicted in parallel. This design further refines generation, yielding a sharp reduction in penetration metrics ($\rm IV_r$ from $6.99$ to $5.48$, $\rm IVU$ from $0.11$ to $0.08$) and boosting physical plausibility to $97.94$. 
\textbf{R5} further incorporates pre-training on GraspXL, achieving the best performance across metrics (\emph{e.g.}, 98.25 in Phy, 5.31 in $\rm IV_r$, and 0.31 in SD).

\subsection{Comparison with state of the art}
\label{subsec:sota}
In this section, we compare our proposed HO-Flow approach with state-of-the-art approaches on three mainstream hand-object motion benchmarks. 

\noindent \textbf{GRAB.} Table~\ref{tab:sota_grab} provides a comprehensive quantitative evaluation on GRAB benchmark. Compared with LatentHOI, HO-Flow substantially reduces unnatural interactions, achieving the lowest interpenetration volume (IV) and intersection depth (ID) for both hands. In particular, HO-Flow improves physical plausibility (Phy) from 96.16\% to 98.25\% and significantly increases semantic diversity (SD), nearly tripling LatentHOI's score. Overall, these results demonstrate that our flow-based approach generates physically grounded and diverse hand-object interactions, preserving realistic contact while effectively suppressing unnatural interpenetrations.

\input{tables/sota_grab}

\input{tables/sota_oakink}

\noindent \textbf{OakInk.} We further evaluate the generalization of HO-Flow model on the out-of-distribution OakInk benchmark, which features a much broader set of unseen objects than prior datasets. As reported in Table~\ref{tab:sota_oakink}, HO-Flow achieves consistently strong performance in both motion quality and diversity. In particular, our method generalizes well to novel object geometries, yielding a clear reduction in hand-object interpenetrations and intersection depths compared with the recent LatentHOI~\cite{Muchen_LatentHOI}. Moreover, HO-Flow improves physical plausibility (Phy) to $89.76\%$, outperforming the text-driven Text2HOI~\cite{cha2024text2hoi} and diffusion-based approaches~\cite{tevet2022human,chen2023executing}. Despite the increased difficulty of interacting with diverse unseen objects, HO-Flow maintains the highest sample diversity (SD), indicating that it captures the underlying distribution of hand-object interactions without sacrificing realism. Notably, in this more challenging setting with more diverse objects, pre-training on GraspXL provides a larger gain in performance, further improving physical fidelity and reducing penetration-related interaction artifacts.

\begin{figure}[t]
  \vspace{-0.3cm}
  \centering  
  \includegraphics[width=1.0\linewidth]{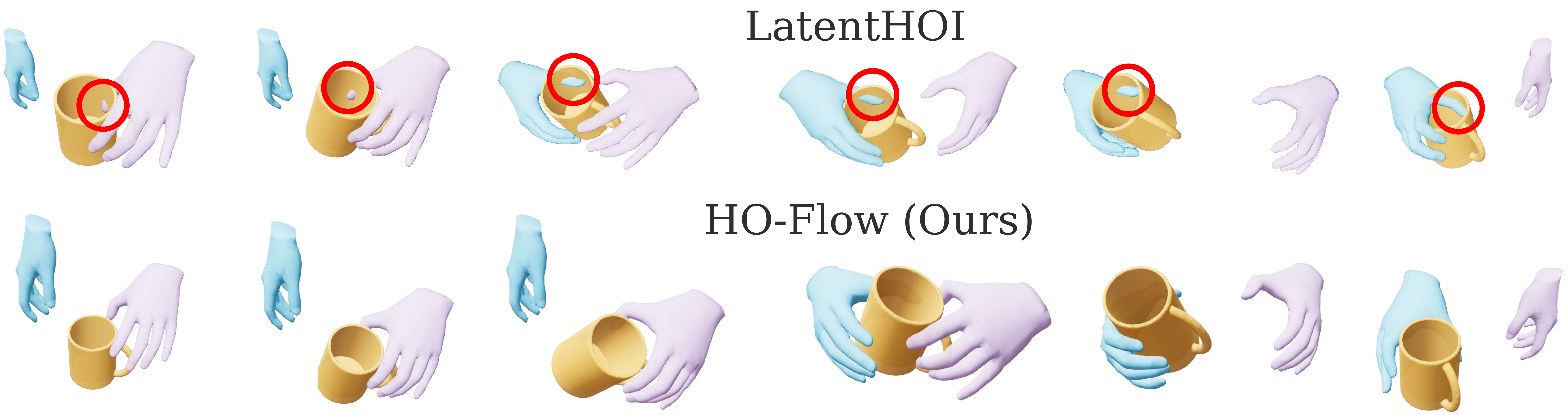}
  \vspace{-0.6cm}
  \caption{Qualitative comparison with the state-of-the-art LatentHOI approach~\cite{Muchen_LatentHOI}. Our HO-Flow model can generate more realistic hand-object interactions with more natural contacts and significantly less penetrations.}
  \label{fig:qcomp}
  \vspace{-0.2cm}
\end{figure}

\input{tables/sota_dexycb}

\input{tables/user}

\noindent \textbf{DexYCB.} Table~\ref{tab:sota_dexycb} further demonstrates the effectiveness of HO-Flow in synthesizing single-hand manipulation sequences on the DexYCB benchmark. Compared to MDM~\cite{tevet2022human} and the state-of-the-art LatentHOI~\cite{Muchen_LatentHOI}, our approach shows a significant reduction in mesh penetration, decreasing the Intersection Depth (ID) from 2.01mm to 1.20mm. Furthermore, HO-Flow achieves a notable leap in physical plausibility (Phy), reaching 95.41\% compared to LatentHOI's 88.52\%, while simultaneously improving semantic diversity (SD). These results confirm that our model not only excels in bimanual manipulation synthesis but also possesses a strong generative capability to produce realistic and diverse interaction sequences for single-hand scenarios.

\begin{figure}[t]
  \centering  
  \includegraphics[width=1.0\linewidth]{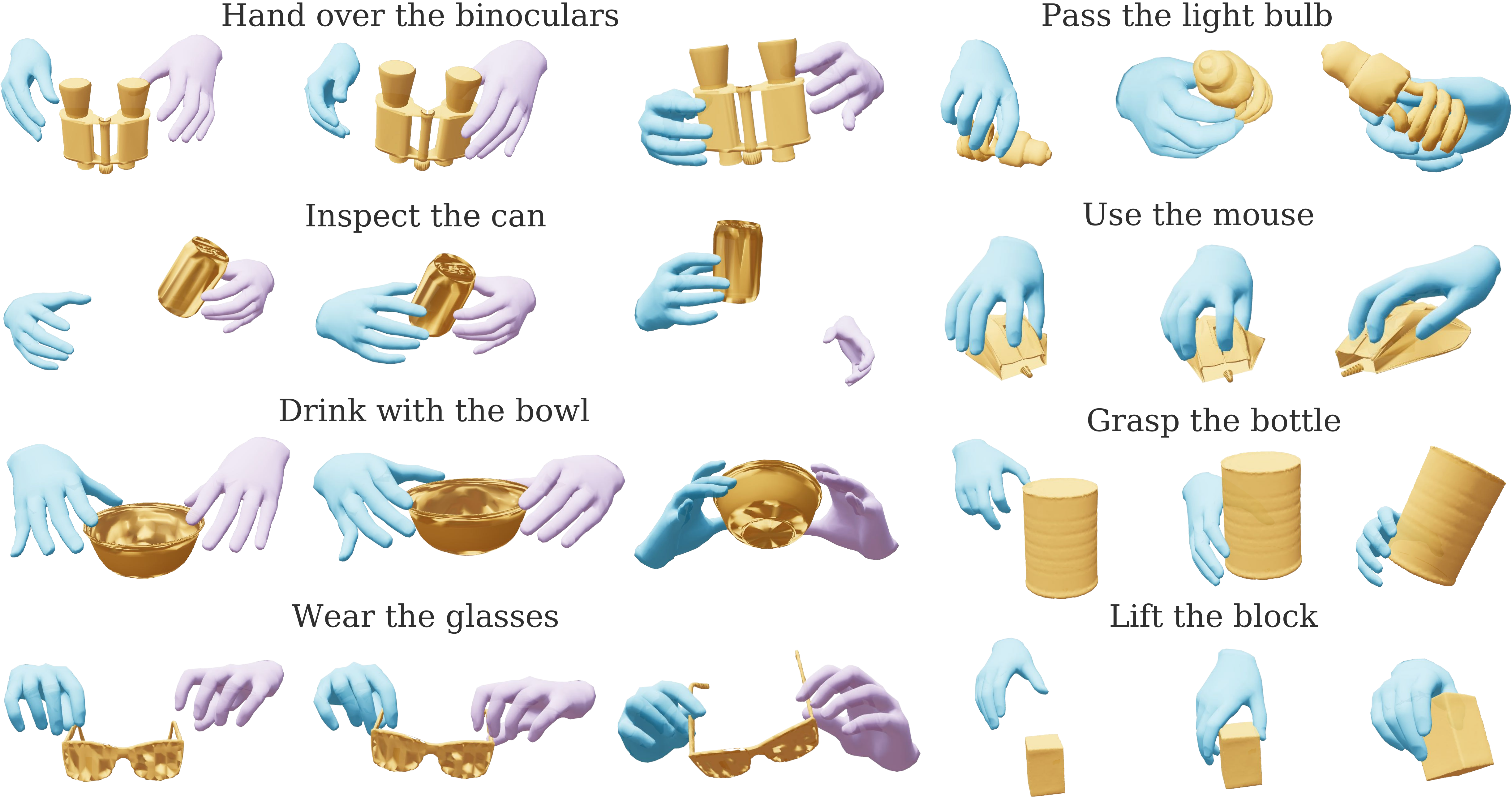}
  \vspace{-0.2cm}
  \caption{Qualitative results of HO-Flow on OakInk and DexYCB benchmarks. Our approach can faithfully synthesize hand and object motions with natural interactions.}
  \label{fig:qres}
  \vspace{-0.1cm}
\end{figure}

\noindent \textbf{Qualitative performance.} Figure~\ref{fig:qcomp} qualitatively compares HO-Flow with the state-of-the-art LatentHOI~\cite{Muchen_LatentHOI}, showing that our model synthesizes more realistic interaction motions with much fewer hand-object penetrations. We also conduct a user study with three human evaluators to compare HO-Flow against LatentHOI~\cite{Muchen_LatentHOI}. For each evaluator, we randomly sample 50 comparison pairs for assessment. For each pair on different benchmarks, the evaluators choose the preferred result based on visual quality and alignment with the task description. As shown in Table~\ref{tab:user}, our method is consistently favored across all benchmarks. Figure~\ref{fig:qres} provides qualitative examples of HO-Flow on objects from OakInk and DexYCB datasets, in addition to those on GRAB shown in Figure~\ref{fig:qcomp}. We observe that HO-Flow synthesizes natural, realistic hand–object interactions in both bi-manual and single-hand settings, across a wide range of objects and tasks.

%% file: tables/abl_vae.tex
\begin{table}[t]
\centering
\caption{Ablation experiments for interaction-aware autoencoder on GRAB dataset. Our proposed model that captures both spatial and temporal interaction features yields the best performance in encoding hand and object motions.}
\vspace{-0.1cm}
\setlength{\tabcolsep}{4pt}
\renewcommand\arraystretch{1.0}
\begin{tabular}{ccccccccccc}
\toprule
\multirow{2}{*}{} &
  \multirow{2}{*}{\begin{tabular}[c]{@{}c@{}}Object\\ kine.\end{tabular}} &
  \multirow{2}{*}{\begin{tabular}[c]{@{}c@{}}Hand\\ kine.\end{tabular}} &
  \multirow{2}{*}{\begin{tabular}[c]{@{}c@{}}Sep.\\ latents\end{tabular}} &
  \multirow{2}{*}{\begin{tabular}[c]{@{}c@{}}Pre-\\ train\end{tabular}} &
  \multirow{2}{*}{$\rm E_v^{r}\downarrow$} &
  \multirow{2}{*}{$\rm E_j^{r}\downarrow$} &
  \multirow{2}{*}{$\rm E_v^{l}\downarrow$} &
  \multirow{2}{*}{$\rm E_j^{l}\downarrow$} &
  \multirow{2}{*}{$\rm E_o\downarrow$} &
  \multirow{2}{*}{$\rm CD_{o}\downarrow$} \\ 
   &  &  &  &  &  &  &  &  &  &  \\ \midrule
R1 &$\times$&$\times$&$\times$&$\times$&23.93&24.34& 19.14&19.55&16.34&1.36  \\
R2 &\checkmark&$\times$&$\times$&$\times$&22.15&22.56&20.51&20.58&8.93&0.71  \\
R3 &\checkmark&\checkmark&$\times$&$\times$&21.69&21.97&16.59&16.90&5.19&0.37  \\
R4 &\checkmark&\checkmark&\checkmark&$\times$&17.29&17.61&  13.05&13.33&3.70&0.26  \\
R5 &\checkmark&\checkmark&\checkmark&\checkmark&\textbf{9.13}&\textbf{9.41}&\textbf{8.03}&\textbf{8.31}&\textbf{2.33}&\textbf{0.12} \\ \bottomrule
\end{tabular}
\label{tab:iv}
\vspace{-0.3cm}
\end{table}

%% file: tables/abl_fm.tex
\begin{table}[!t]
\centering
\footnotesize
\caption{Ablation experiments for the auto-regressive flow matching model on GRAB dataset. Our approach achieves the best performance over other variants. Seq. indicates whether the model takes a temporal sequence as input.}
\vspace{-0.2cm}
\setlength{\tabcolsep}{1.2pt} 
\renewcommand\arraystretch{1.0}
\begin{tabular}{l cc ccccccccccc}
\toprule
\multirow{2}{*}{} & \multicolumn{2}{c}{Representation} & \multirow{2}{*}{\begin{tabular}[c]{@{}c@{}}Auto-\\ reg.\end{tabular}} & \multirow{2}{*}{\begin{tabular}[c]{@{}c@{}}Pre-\\ train\end{tabular}} & \multirow{2}{*}{$\rm IV_{r}\downarrow$} & \multirow{2}{*}{$\rm IV_{l}\downarrow$} & \multirow{2}{*}{$\rm ID_{r}\downarrow$} & \multirow{2}{*}{$\rm ID_{l}\downarrow$} & \multirow{2}{*}{$\rm CR_{r}\uparrow$} & \multirow{2}{*}{$\rm CR_{l}\uparrow$} & \multirow{2}{*}{$\rm IVU\downarrow$} & \multirow{2}{*}{$\rm Phy\uparrow$} & \multirow{2}{*}{$\rm SD\uparrow$} \\
\cmidrule(lr){2-3}
 & Latent & Seq. & & & & & & & & & & & \\ \midrule
R1 & $\times$ & \checkmark & $\times$ & $\times$ & 10.35 & 2.67 & 1.14 & 0.48 & 11.68 & 3.73 & 0.13 & 84.89 & 0.24 \\ 
R2 & \checkmark& $\times$  & $\times$ & $\times$ & 8.13 & 1.94 & 0.98 & 0.43 & \textbf{12.54} & 0.71 & 0.13 & 88.76 & 0.12 \\
R3 & \checkmark & \checkmark & $\times$ & $\times$ & 6.99 & 1.82 & 0.82 & 0.36 & 10.27 & 3.07 & 0.11 & 92.69 & 0.26 \\
R4 & \checkmark & \checkmark & \checkmark & $\times$ & 5.48 & 1.35 & 0.63 & 0.30 & 11.26 & 2.86 & 0.08 & 97.94 & 0.30 \\
R5 & \checkmark & \checkmark & \checkmark & \checkmark & \textbf{5.31} & \textbf{1.24} & \textbf{0.61} & \textbf{0.26} & 11.38 & \textbf{3.96} & \textbf{0.07} & \textbf{98.25} & \textbf{0.31} \\ 
\bottomrule
\end{tabular}
\label{tab:fm}
\vspace{-0.3cm}
\end{table}

%% file: tables/sota_grab.tex
\begin{table}[t]
\centering
\footnotesize
\caption{Comparison with state-of-the-art methods on GRAB benchmark. HO-Flow demonstrates strong generalization ability to faithfully interact with diverse objects. pt denotes whether this model has been pre-trained on GraspXL data.}
\vspace{-0.2cm}
\setlength{\tabcolsep}{3pt}
\renewcommand\arraystretch{1.1}
\begin{tabular}{lccccccccc}
\toprule
Methods & $\rm IV_{r}\downarrow$ & $\rm IV_{l}\downarrow$ & $\rm ID_{r}\downarrow$ & $\rm ID_{l}\downarrow$ & $\rm CR_{r}\uparrow$ & $\rm CR_{l}\uparrow$ & $\rm IVU\downarrow$ & $\rm Phy\uparrow$ & $\rm SD\uparrow$ \\ 
\midrule
IMoS~\cite{ghosh2023imos} & 10.38 & - & 1.25 & - & 4.61 & - & 0.53 & 84.88 & 0.00 \\
MDM~\cite{tevet2022human} & 9.12 & 2.61 & 1.24 & 0.51 & 8.21 & 1.29 & 0.19 & 89.81 & 0.18 \\
MLD~\cite{chen2023executing} & 9.62 & 3.14 & 1.06 & 0.49 & 10.23 & 0.87 & 0.24 & 85.68 & 0.20 \\
LatentHOI~\cite{Muchen_LatentHOI} & 6.38 & 1.66 & 0.77 & 0.29 & \textbf{11.94} & 1.11 & 0.10 & 96.16 & 0.13  \\
\midrule
HO-Flow, wo pt &5.48&1.35&0.63&0.30&11.26&2.86&0.08&97.94&0.30\\
HO-Flow, w pt &\textbf{5.31}&\textbf{1.24}&\textbf{0.61}&\textbf{0.26}&11.38&\textbf{3.96}&\textbf{0.07}&\textbf{98.25}&\textbf{0.31}\\
\bottomrule
\end{tabular}
\label{tab:sota_grab}
\end{table}

%% file: tables/sota_oakink.tex
\begin{table}[!t]
\centering
\footnotesize
\caption{Comparison with state-of-the-art methods on OakInk benchmark. HO-Flow achieves superior performance in terms of both motion quality and diversity. pt denotes whether this model has been pre-trained on GraspXL data.}
\vspace{-0.2cm}
\setlength{\tabcolsep}{2.4pt}
\renewcommand\arraystretch{1.1}
\begin{tabular}{lccccccccc}
\toprule
Methods & $\rm IV_{r}\downarrow$ & $\rm IV_{l}\downarrow$ & $\rm ID_{r}\downarrow$ & $\rm ID_{l}\downarrow$ & $\rm CR_{r}\uparrow$ & $\rm CR_{l}\uparrow$ & $\rm IVU\downarrow$ & $\rm Phy\uparrow$ & $\rm SD\uparrow$ \\ 
\midrule
Text2HOI~\cite{cha2024text2hoi}& 15.19 & 11.54 & 2.14 & 1.39 & \textbf{11.24} & \textbf{6.33} & 0.26 & 82.58 & 0.21 \\
MDM~\cite{tevet2022human}& 8.46 & 2.47 & 1.69 & 0.34 & 4.97 & 1.02 & 0.20 & 60.89 & 0.23 \\
MLD~\cite{chen2023executing}& 9.15 & 4.25 & 1.79 & 0.56 & 5.36 & 0.77 & 0.29 & 46.41 & 0.32 \\
LatentHOI~\cite{Muchen_LatentHOI}& 7.22 & 3.11 & 1.10 & 0.37 & 7.80 & 1.73 & 0.14 & 71.24 & 0.22 \\
\midrule
HO-Flow, wo pt &5.82&2.05&0.66&0.30&8.76&2.34&0.11&83.62&0.31\\
HO-Flow, w pt & \textbf{4.10} & \textbf{1.72} & \textbf{0.45} & \textbf{0.26} & 8.85 & 3.78 & \textbf{0.09} & \textbf{89.76} & \textbf{0.33}\\
\bottomrule
\end{tabular}
\label{tab:sota_oakink}
\vspace{-0.2cm}
\end{table}

%% file: tables/sota_dexycb.tex
\begin{table}[t]
\centering
\footnotesize
\caption{Comparison with state-of-the-art methods on DexYCB benchmark. HO-Flow also shows strong performance for synthesizing single-hand manipulation sequences. pt denotes whether this model has been pre-trained on GraspXL data.}
\vspace{-0.2cm}
\setlength{\tabcolsep}{8pt}
\renewcommand\arraystretch{1.1}
\begin{tabular}{lcccccc}
\toprule
Methods & $\rm IV\downarrow$ & $\rm ID\downarrow$ & $\rm CR_{r}\uparrow$ & $\rm IVU\downarrow$ & $\rm Phy\uparrow$ & $\rm SD\uparrow$ \\ 
\midrule
MDM~\cite{tevet2022human} & 7.78 & 2.10 & 8.87 & 0.12 & 86.22 & 0.13 \\
LatentHOI~\cite{Muchen_LatentHOI} & 7.70 & 2.01 & 11.98 & 0.13 & 88.52 & 0.13  \\
\midrule
HO-Flow, wo pt &6.84&1.82&12.06&0.11&90.77&0.20\\
HO-Flow, w pt &\textbf{6.37}&\textbf{1.20}&\textbf{13.83}&\textbf{0.10}&\textbf{95.41}&\textbf{0.20}\\
\bottomrule
\end{tabular}
\label{tab:sota_dexycb}
\vspace{-0.1cm}
\end{table}

%% file: tables/user.tex
\begin{table}[!t]
\centering
\footnotesize
\caption{Qualitative user study comparing HO-Flow with LatentHOI~\cite{Muchen_LatentHOI} across three benchmarks. Values indicate the preference ratio (higher is better).}
\vspace{-0.2cm}
\setlength{\tabcolsep}{10pt}
\renewcommand\arraystretch{1.1}
\begin{tabular}{lccc}
\toprule
Methods& GRAB & OakInk & DexYCB\\ 
\midrule
LatentHOI~\cite{Muchen_LatentHOI} & 0.37 & 0.28 & 0.38 \\
HO-Flow (Ours)&\textbf{0.63}&\textbf{0.72}&\textbf{0.62}\\
\bottomrule
\end{tabular}
\label{tab:user}
\vspace{-0.5cm}
\end{table}

%% file: sections/conclusion.tex
\section{Conclusion}
This work presents HO-Flow, a novel framework for generating realistic 3D hand-object interaction motion sequences. It aims to improve physical plausibility, long-horizon temporal coherence, and generalization across diverse actions and objects. HO-Flow combines an expressive interaction-aware motion representation with efficient hand-object interaction synthesis in the continuous space. Inter-VAE encodes short-horizon hand-object interaction sequences into unified latent codes, capturing both global motion trajectories and fine-grained, contact-rich coordination through hand-centric geometric cues. On top of this representation, a masked flow matching model generates motion latent tokens with auto-regressive temporal reasoning, reducing jitter, outliers, and interpenetration. Predicting object motion relative to the initial frame further improves robustness to dataset-specific coordinate conventions and supports large-scale pre-training on synthetic data. Experiments on GRAB, OakInk, and DexYCB show that HO-Flow advances the state of the art in terms of both physical plausibility and motion diversity for text-conditioned hand-object motion generation.

%% file: appendix.tex
In appendix, we provide implementation details about our network architecture in Section~\ref{supmat:net}. Section~\ref{supmat:train} further describes the training and testing procedures of our model. Finally, Section~\ref{supmat:eval} discusses details about evaluation protocols and additional experimental results.

\section{Network Architecture}
\label{supmat:net}

\subsection{Interaction-aware variational autoencoder}
Inter-VAE is implemented as a shared spatial encoder followed by two branch-specific temporal VAEs for object and hand motion. For each frame, we build a 57-dimensional point-wise interaction descriptor by concatenating the canonical object coordinates, world-frame object coordinates, root-relative object coordinates, and the object coordinates expressed in the 16 local MANO~\cite{MANO:SIGGRAPHASIA:2017} joint frames. These features are processed with three set-abstraction layers~\cite{qi2017pointnet++} and a final global grouping layer to produce a shared frame-level feature. The shared spatial feature is then concatenated with branch-specific motion descriptors: a 9D object pose vector, consisting of 6D rotation~\cite{zhou2019continuity} and translation, for the object branch, and a 246D hand-object descriptor for the hand branch, consisting of 16 joint 6D rotations, hand translation, root-relative object translation, and per-joint object rotations and translations expressed in local hand coordinates.

Both branches adopt the same temporal VAE design, comprising an input 1D convolution, two stride-2 downsampling stages with dilated 1D convolution blocks, and $1\times1$ convolutions to predict $\boldsymbol{\mu}$ and $\log \boldsymbol{\sigma}^{2}$ for reparameterization~\cite{kingma2013auto}. The temporal decoders mirror this architecture with linear upsampling and residual temporal blocks. The object decoder reconstructs a single 9D object trajectory, while the hand decoder predicts $16\times6+3$ motion parameters conditioned on the concatenated latent code $[\mathbf{z}_h, \mathbf{z}_o]$. Importantly, $\mathbf{z}_o$ is detached when conditioning the hand decoder, preventing gradients from the hand reconstruction loss from propagating into the object branch and thereby reducing conflicts between optimizing $\mathbf{z}_h$ and $\mathbf{z}_o$ during training.

\subsection{Auto-regressive flow matching}

In implementation, each motion token $\mathbf{z}_t \in \mathbb{R}^{128}$ is first projected to a 1024-dimensional hidden space and augmented with sinusoidal positional encodings before being processed by the masked auto-regressive transformer. This compact transformer backbone is adopted, with adaptive LayerNorm conditioning, a hidden dimension of 1024, 16 attention heads, an MLP expansion dimension of 4096, and a dropout rate of 0.2. The text and geometry conditions are encoded separately and projected into the same 1024-dimensional space, after which they are summed to form the final conditioning vector. Specifically, we use a frozen CLIP~\cite{radford2021learning} text embedding of dimension 512 and a BPS~\cite{prokudin2019efficient} object descriptor of
\vspace{-0.8cm}
\input{tables/algo}
\noindent dimension 4096, each followed by a linear projection. The transformer blocks are modulated by this condition through AdaLN, and the modulation layers are zero-initialized at the beginning of training for stable optimization.

For token prediction, we employ a flow matching model built on a lightweight conditional MLP following a SiT-style architecture~\cite{ma2024sit}, comprising 16 residual AdaLN~\cite{peebles2023scalable} blocks with width 1792. The model is conditioned on both the contextual feature $\mathbf{h}_t$ from the masked auto-regressive transformer and the continuous flow time $\tau$ via dedicated timestep embeddings. Training operates exclusively on masked tokens: under a cosine mask-ratio schedule, masked positions are corrupted with either a learnable \texttt{[MASK]} token or Gaussian noise, while unmasked tokens supply temporal context for auto-regressive inpainting. At inference, generation begins from a fully masked latent sequence and progressively unmasks tokens according to a random priority order consistent across iterations, preventing repeated overwriting of already confident predictions. At each iteration, the model predicts currently masked tokens with classifier-free guidance~\cite{ho2022classifier}, and the final synthesized latent sequence is decoded by the frozen Inter-VAE decoder into hand-object motion trajectories.

\section{Training and Testing Details}
\label{supmat:train}
\subsection{Training details}
We pre-train both the interaction-aware variational autoencoder and the context-aware auto-regressive flow matching model on more than 5 million synthetic motion sequences from the GraspXL dataset~\cite{zhang2024graspxl}. Each motion sequence contains only the right hand interacting with an object and consists of 155 motion frames. All sequences are padded to 160 frames by repeating the last original frame. For training the auto-regressive transformer, different from Equation~5 in the main paper, we construct the latent token $\mathbf{z}_t$ as
\begin{equation}
\begin{split}
\mathbf{z}_t &= \left[\mathbf{z}^{r}_{o,t}\;\Vert\;\mathbf{z}^{r}_{h,t}\;\Vert\;\mathbf{z}^{r}_{o,t}\;\Vert\;\mathbf{z}^{r}_{h,t}\right]\in\mathbb{R}^{128},
\end{split}
\label{eq:token_def}
\end{equation}
where the right-hand components are duplicated to represent the left-hand counterparts. In this way, single-hand manipulation motions are treated as symmetric bi-manual hand-object motions, which is compatible with subsequent model fine-tuning on real bi-manual motion data. We fine-tune the pre-trained checkpoints of our VAE and flow matching models for 100k and 300k steps, respectively, using a learning rate of $1\times10^{-4}$ and decay to $1\times10^{-5}$ in training.

\subsection{Testing details}
\noindent Algorithm~\ref{algo} summarizes the motion generation process of our HO-Flow approach. Given a task description and an object point cloud, the model first encodes the task semantics using the CLIP text encoder~\cite{radford2021learning} and extracts the object representation using BPS~\cite{prokudin2019efficient}, after which the two features are projected and fused into a single condition embedding. Conditioned on this embedding, HO-Flow then generates the motion latent sequence auto-regressively: at each time step, the masked auto-regressive transformer takes the partially generated sequence as input, produces the contextual feature for the current token, and uses it to guide the flow matching model, which transforms an initial Gaussian noise sample into the target latent token by integrating the learned velocity field from $\tau=0$ to $\tau=1$. After all latent tokens are generated, the full latent sequence is split into hand- and object-related components and passed to the frozen decoders, where the object decoder first reconstructs the object motion trajectory and the hand decoder subsequently predicts the hand motion conditioned on both the generated latents and the decoded object motion, yielding the final synthesized hand motion $\hat{\mathbf{T}}_h$ and object motion $\hat{\mathbf{T}}_o$.

\section{Experimental Results}
\label{supmat:eval}
\subsection{Evaluation metrics}

\noindent \textbf{Hand joint error} ($\rm E_{j}$). We evaluate 3D hand reconstruction performance by computing the average $\ell_2$ distance between the reconstructed MANO hand joints $\rm \hat{\mathbf{j}}_{h}\in\mathbb{R}^{N\times21\times3}$ and their ground-truth counterparts $\rm \mathbf{j}_{h}$.

\noindent \textbf{Hand vertex error} ($\rm E_{v}$). Similarly, we compute $\rm E_{v}$ as the average $\ell_2$ distance between the predicted vertices $\rm \hat{\mathbf{v}}_{h}\in\mathbb{R}^{N\times778\times3}$ and its ground truth $\mathbf{v}_h$.

\noindent \textbf{Object translation error} ($\rm E_{o}$). We compute $\rm E_{o}$ as the average $\ell_2$ distance between the predicted object translation $\rm \hat{\mathbf{t}}_{o}\in\mathbb{R}^{N\times3}$ and its ground-truth value.

\noindent \textbf{Object chamfer distance} ($\rm CD_{o}$). Directly comparing object rotations is often ill-posed because different object categories may exhibit different symmetries. We therefore use the chamfer distance to account for orientation error:
\begin{equation}
{\rm CD_{o}}({\hat{\mathbf{p}}_{o}^{w}}, \mathbf{p}_{o}^{w}) = \frac{1}{|\hat{\mathbf{p}}_{o}^{w}|} \sum_{p \in \hat{\mathbf{p}}_{o}^{w}} \min_{q \in {\mathbf{p}}_{o}^{w}} \|p - q\|_2^2 + \frac{1}{|{\mathbf{p}}_{o}^{w}|} \sum_{q \in {\mathbf{p}}_{o}^{w}} \min_{p \in \hat{\mathbf{p}}_{o}^{w}} \|q - p\|_2^2,
\end{equation}
where $\hat{\mathbf{p}}_{o}^{w}\in\mathbb{R}^{N\times1024\times3}$ and $\mathbf{p}_{o}^{w}\in\mathbb{R}^{N\times1024\times3}$ denote the reconstructed and ground-truth object point clouds in the world coordinate frame, respectively.

\noindent \textbf{Contact ratio} ($\rm CR$). Following previous work~\cite{Muchen_LatentHOI}, we define $\rm CR$ as the proportion of hand mesh vertices that are in contact with the object mesh. A hand vertex is considered to be in contact if its signed distance to the object mesh is less than 0.45\,mm. In practice, we compute this metric separately for the right and left hands, denoted by $\rm CR_{r}$ and $\rm CR_{l}$, respectively.

\noindent \textbf{Interpenetration volume per contact unit} ($\rm IVU$). We follow the previous work~\cite{Muchen_LatentHOI} to compute $\rm IVU$ to quantify the extent of penetration relative to the degree of contact:
\begin{equation}
{\rm IVU} = \frac{{\rm IV_{r}} + {\rm IV_{l}}}{{\rm CR_{r}} \times {\rm M_{r}} + {\rm CR_{l}} \times {\rm M_{l}}},
\end{equation}
where $\rm IV$ denotes the interpenetration volume, $\rm CR$ denotes the contact ratio, and $\rm M$ denotes the hand mesh surface area.

\noindent \textbf{Physical plausibility} ($\rm Phy$). The Phy metric heuristically assesses the plausibility of a grasp under the principle that the contact forces must be able to support the object once it is no longer in contact with the ground:
\begin{equation}
\mathrm{Phy} = \frac{1}{N}\sum_{i=1}^{N} \mathds{1}\!\left(\mathrm{CR}_{r}^{\,i} > 1\% \text{ or } \mathrm{CR}_{l}^{\,i} > 1\%\right).
\end{equation}

\noindent \textbf{Sample diversity} ($\rm SD$). Following prior work~\cite{christen2024diffh2o,Muchen_LatentHOI}, we measure sample diversity as the average pairwise squared $\ell_2$ distance between hand motion sequences:
\begin{equation}
\mathrm{SD}
=
\frac{2}{N(N-1)}
\sum_{i=1}^{N}
\sum_{j=i}^{N}
\left\|\mathbf{v}_{h}^{i} - \mathbf{v}_{h}^{j}\right\|_2^2
\end{equation}

\input{tables/supmat_diffusion}

\subsection{Diffusion vs. Flow Matching}
We compare two alternative generative training paradigms for HO-Flow on the GRAB dataset: the standard diffusion formulation using DDPM~\cite{ho2020ddpm} and the flow matching formulation~\cite{lipman2023flowmatching}. As shown in Table~\ref{tab:supmat_diff}, flow matching consistently outperforms diffusion across different evaluation metrics. In particular, it achieves lower interpenetration volume and distance for both right and left hands ($\rm IV_r$, $\rm IV_l$, $\rm ID_r$, and $\rm ID_l$), while also improving contact ratio ($\rm CR_r$ and $\rm CR_l$). Moreover, flow matching yields better physical plausibility, reflected by reduced $\rm IVU$ and improved $\rm Phy$, while maintaining slightly higher diversity ($\rm SD$). These results suggest that flow matching provides a more effective and stable learning objective for modeling hand-object interactions, and we therefore adopt it as the default training strategy in our HO-Flow approach. 

\input{tables/supmat_sampling}

\begin{figure}[ht]
  \centering  
  \includegraphics[width=0.88\linewidth]{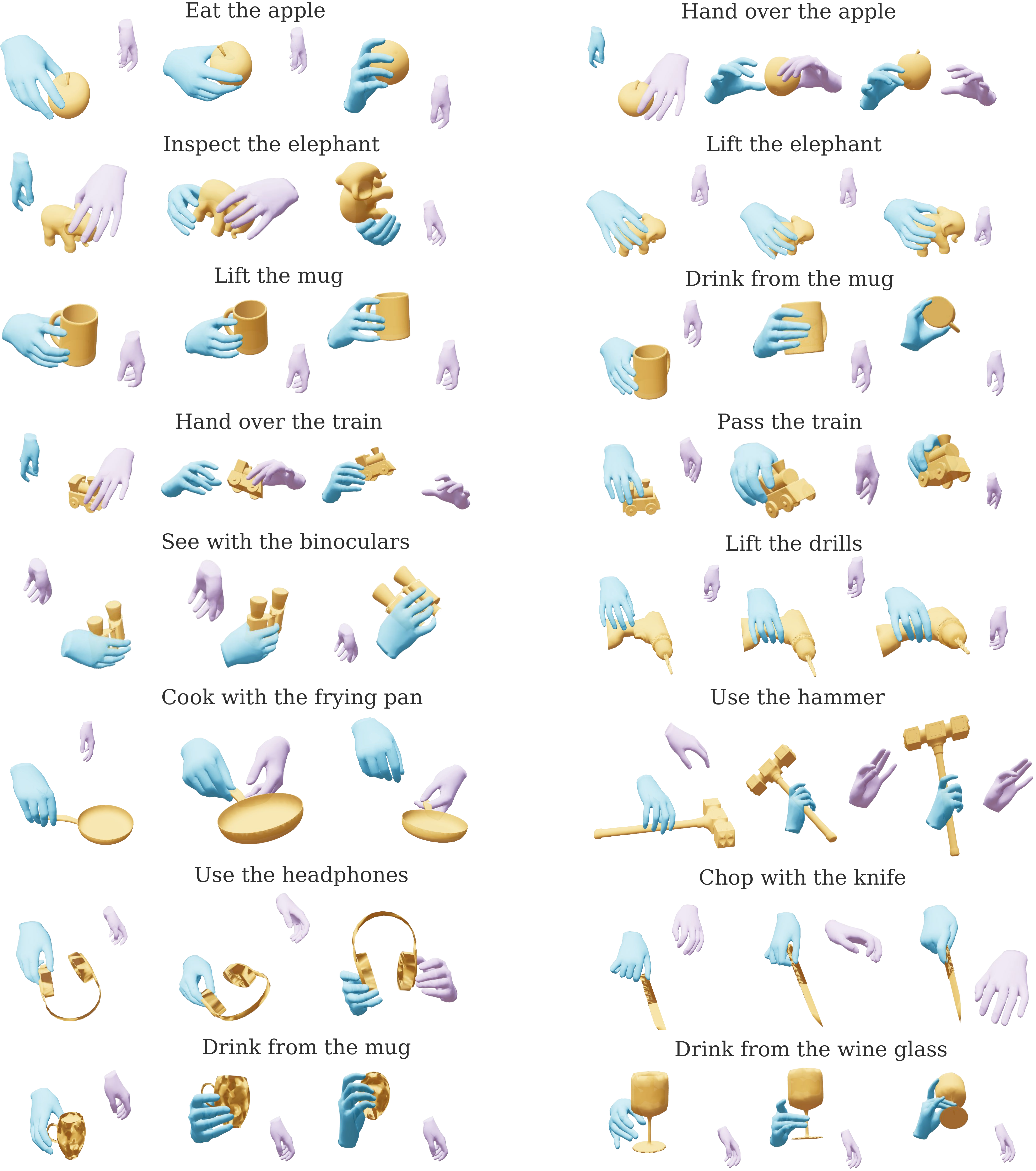}
  \vspace{-0.2cm}
  \caption{Qualitative results of proposed HO-Flow approach on GRAB (first four rows) and OakInk (last four rows) benchmarks. Our approach can synthesize realistic hand and object interaction motions across diverse objects and tasks.}
  \label{fig:supmat_demo}
  \vspace{-0.1cm}
\end{figure}

\subsection{Impact of prediction steps}
Table~\ref{tab:supmat_step} presents an ablation study on the number of prediction steps at test time for HO-Flow on the GRAB dataset. Overall, increasing the number of steps allows motion tokens to be revealed more gradually, improving motion quality and physical plausibility at the cost of higher inference latency. As shown in Table~\ref{tab:supmat_step}, increasing the number of steps from 6 to 18 reduces $\rm IV_r$ from 8.51 $\rm cm^3$ to 5.31 $\rm cm^3$, reduces $\rm ID_r$ from 0.77 $\rm cm$ to 0.61 $\rm cm$, increases $\rm CR_l$ from 3.29 to 3.96, lowers $\rm IVU$ from 0.13 to 0.07, and improves $\rm Phy$ from 90.63 to 98.25. Although using 24 steps yields slight gains on a few metrics, the improvement over 18 steps is marginal compared with the increased latency, which rises from 6.89s to 9.26s to generate each motion sequence. Therefore, we adopt 18 steps as a good trade-off between generation quality and efficiency.

\begin{figure}[t]
  \centering  
  \includegraphics[width=0.95\linewidth]{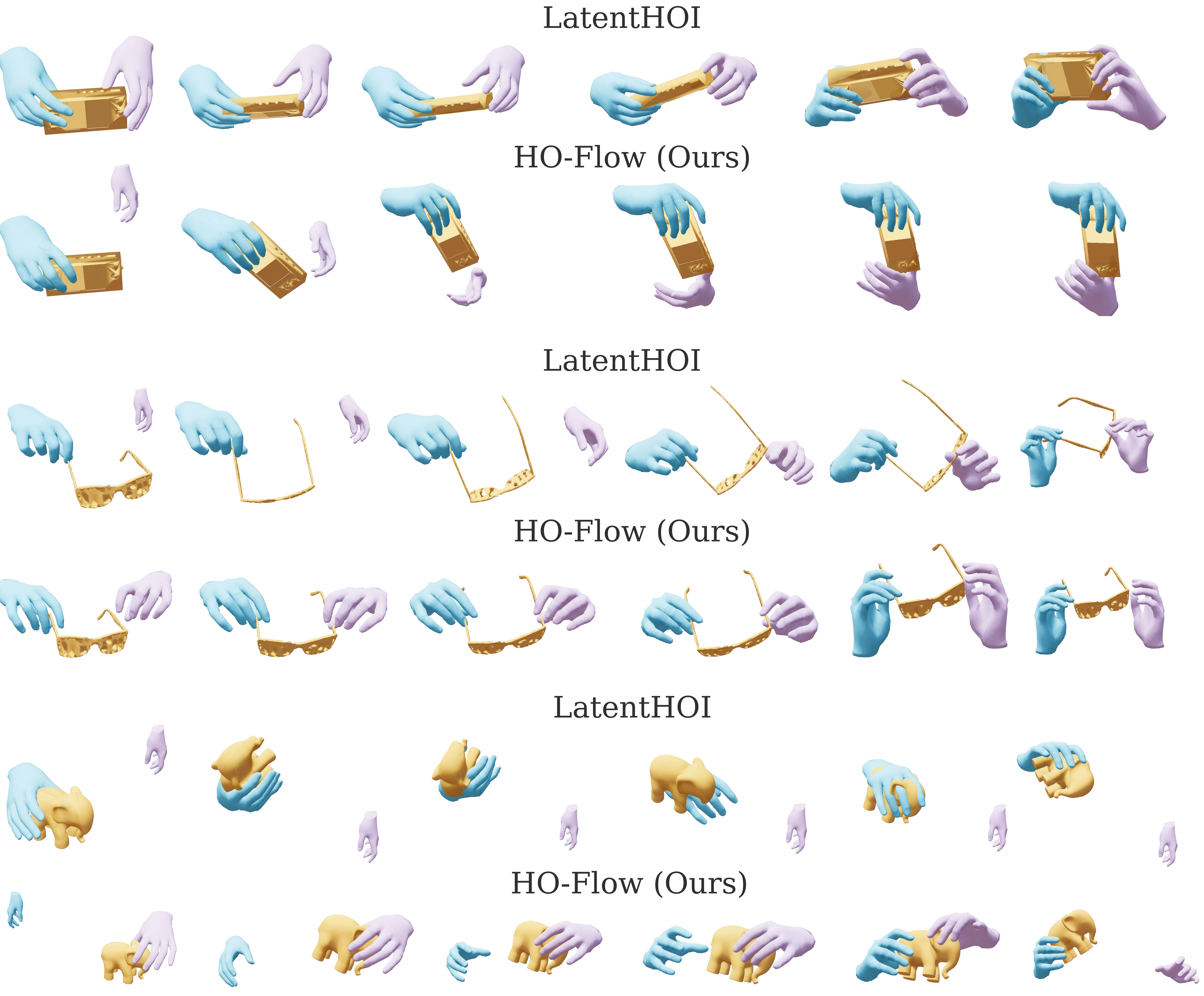}
  \vspace{-0.3cm}
  \caption{Qualitative comparison with the state-of-the-art LatentHOI approach~\cite{Muchen_LatentHOI} on GRAB and OakInk benchmarks. }
  \label{fig:supmat_qcomp}
  \vspace{-0.1cm}
\end{figure}

\begin{figure}[!t]
  \centering  
  \includegraphics[width=0.95\linewidth]{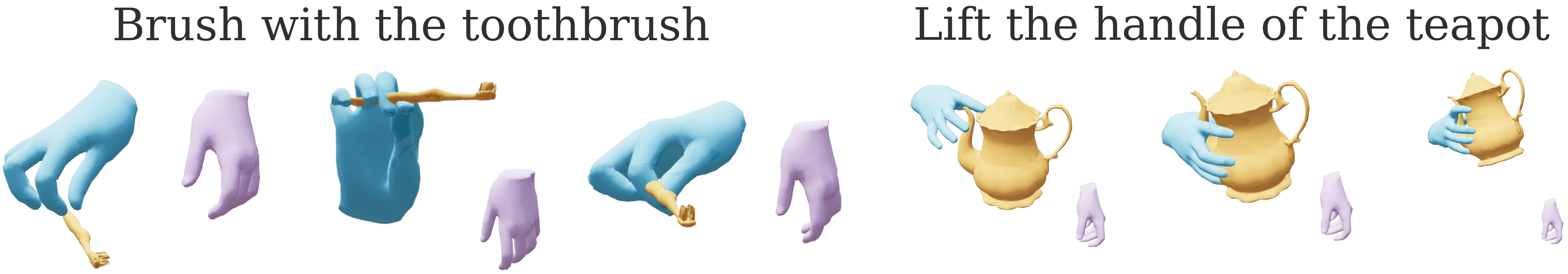}
  \vspace{-0.3cm}
  \caption{Failure cases of our HO-Flow approach on OakInk objects.}
  \label{fig:supmat_failure}
  \vspace{-0.4cm}
\end{figure}

\subsection{Additional qualitative results}
Figure~\ref{fig:supmat_demo} presents qualitative results of our HO-Flow model on diverse unseen objects from the GRAB~\cite{GRAB:2020} and OakInk~\cite{yang2022oakink} datasets. The first four rows and the last four rows show results from GRAB and OakInk, respectively. The results demonstrate that our model effectively generalizes across a wide range of objects, and produces realistic and natural hand-object interaction motions. Figure~\ref{fig:supmat_qcomp} additionally presents a qualitative comparison with LatentHOI~\cite{Muchen_LatentHOI}, showing that our approach produces more realistic results with fewer penetrations.

Figure~\ref{fig:supmat_failure} presents representative failure cases of our HO-Flow approach on OakInk objects. In the left example, the predicted orientation of the toothbrush is inaccurate, which may further result in noticeable interpenetration between the hand and the object. In the right example, the generated hand fails to grasp the semantically specified part of the teapot. These failure cases suggest that future improvements may come from scaling training with more diverse language instructions and incorporating object affordance cues into the generation of more realistic hand-object interaction motions.

%% file: tables/algo.tex
\begin{algorithm}[H]
\caption{Inference procedure of proposed HO-Flow model}
\begin{algorithmic}
\STATE \textbf{Input:} task description $s$, object point cloud $\mathbf{p}_o$, trained auto-regressive transformer $\mathrm{MAR}$, flow matching model $v_{\phi}$, projection matrices $\mathbf{W}_{\text{text}}$ and $\mathbf{W}_{\text{bps}}$, frozen hand decoder $D_h$, frozen object decoder $D_o$
\STATE \textbf{Output:} synthesized hand motion $\hat{\mathbf{T}}_h$ and object motion $\hat{\mathbf{T}}_o$

\STATE
\begin{tcolorbox}[colback=red!5!white,colframe=red!50!black!50!,left=2pt,right=2pt,top=0pt,bottom=1pt,colbacktitle=red!25!white,title=\textbf{\color{black} Condition Encoding}]
\STATE Encode the task description: $\mathbf{e}_{\text{text}} \leftarrow \mathrm{CLIP}(s)$
\STATE Compute the object representation: $\mathbf{e}_{\text{bps}} \leftarrow \mathrm{BPS}(\mathbf{p}_o)$
\STATE Construct the condition embedding:
\STATE $\mathbf{c} \leftarrow \mathbf{W}_{\text{text}}\mathbf{e}_{\text{text}} + \mathbf{W}_{\text{bps}}\mathbf{e}_{\text{bps}}$
\end{tcolorbox}

\STATE
\begin{tcolorbox}[colback=yellow!5!white,colframe=yellow!50!black!50!,left=2pt,right=2pt,top=0pt,bottom=1pt,colbacktitle=yellow!25!white,title=\textbf{\color{black} Auto-regressive Latent Generation}]
\STATE Initialize the latent sequence $\hat{\mathbf{Z}}=[\,]$
\FOR{$t=1$ to $L$}
    \STATE Form the partially masked sequence $\widetilde{\mathbf{Z}}$ from previously generated tokens and a mask at position $t$
    \STATE Extract contextual features:
    \STATE $\mathbf{H} \leftarrow \mathrm{MAR}(\widetilde{\mathbf{Z}}, \mathbf{c})$
    \STATE Select the context feature of the current token: $\mathbf{h}_t \leftarrow \mathbf{H}[t]$
    \STATE Sample initial noise: $\mathbf{x}_0 \sim \mathcal{N}(\mathbf{0}, \mathbf{I})$
    \STATE Solve the ODE from $\tau=0$ to $\tau=1$ using the velocity field $v_{\phi}(\mathbf{x}_{\tau}, \tau; \mathbf{h}_t)$
    \STATE Obtain the generated token: $\hat{\mathbf{z}}_t \leftarrow \mathbf{x}_{\tau=1}$
    \STATE Append $\hat{\mathbf{z}}_t$ to $\hat{\mathbf{Z}}$
\ENDFOR
\end{tcolorbox}

\STATE
\begin{tcolorbox}[colback=teal!5!white,colframe=teal!50!black!50!,left=2pt,right=2pt,top=0pt,bottom=1pt,colbacktitle=teal!25!white,title=\textbf{\color{black} Motion Decoding}]
\STATE Split each generated token as
\STATE $\hat{\mathbf{z}}_t = [\hat{\mathbf{z}}_{o,t}^{r}\;\Vert\;\hat{\mathbf{z}}_{h,t}^{r}\;\Vert\;\hat{\mathbf{z}}_{o,t}^{l}\;\Vert\;\hat{\mathbf{z}}_{h,t}^{l}]$
\STATE Assemble the generated latent sequence $\hat{\mathbf{Z}}$
\STATE Decode object motion from object latents:
\STATE $\hat{\mathbf{T}}_o \leftarrow D_o(\hat{\mathbf{Z}})$
\STATE Decode right-hand motion from hand latents:
\STATE $\hat{\mathbf{T}}_h^{r} \leftarrow D_h(\hat{\mathbf{Z}}, \hat{\mathbf{T}}_o)$
\STATE Mirror the decoded right-hand motion for left-hand cases:
\STATE $\hat{\mathbf{T}}_h^{l} \leftarrow \mathrm{Mirror}(\hat{\mathbf{T}}_h^{r})$
\STATE Obtain the final hand motion $\hat{\mathbf{T}}_h$ from $\hat{\mathbf{T}}_h^{r}$ and $\hat{\mathbf{T}}_h^{l}$
\STATE Return $\hat{\mathbf{T}}_h$ and $\hat{\mathbf{T}}_o$
\end{tcolorbox}

\end{algorithmic}
\label{algo}
\end{algorithm}
\vspace{-0.5cm}

%% file: tables/supmat_diffusion.tex
\begin{table}[t]
\centering
\footnotesize
\caption{Ablation study of using either diffusion or flow matching approaches to train HO-Flow on GRAB dataset. }
\vspace{-0.2cm}
\setlength{\tabcolsep}{3pt}
\renewcommand\arraystretch{1.1}
\begin{tabular}{ccccccccccc}
\toprule
Approach& $\rm IV_{r}\downarrow$ & $\rm IV_{l}\downarrow$ & $\rm ID_{r}\downarrow$ & $\rm ID_{l}\downarrow$ & $\rm CR_{r}\uparrow$ & $\rm CR_{l}\uparrow$ & $\rm IVU\downarrow$ & $\rm Phy\uparrow$ & $\rm SD\uparrow$ \\ 
\midrule
DDPM~\cite{ho2020ddpm}&6.56&2.09&0.78&0.31&10.53&3.80&0.09&96.00&0.30\\
Flow Matching~\cite{lipman2023flowmatching}&\textbf{5.31}&\textbf{1.24}&\textbf{0.61}&\textbf{0.26}&\textbf{11.38}&\textbf{3.96}&\textbf{0.07}&\textbf{98.25}&\textbf{0.31}\\
\bottomrule
\end{tabular}
\label{tab:supmat_diff}
\end{table}

%% file: tables/supmat_sampling.tex
\begin{table}[t]
\centering
\footnotesize
\caption{Ablation study of prediction steps at test time on GRAB dataset. Latency measures the inference latency for generating a single motion sequence.}
\vspace{-0.2cm}
\setlength{\tabcolsep}{3pt}
\renewcommand\arraystretch{1.1}
\begin{tabular}{cccccccccccc}
\toprule
Steps&Latency& $\rm IV_{r}\downarrow$ & $\rm IV_{l}\downarrow$ & $\rm ID_{r}\downarrow$ & $\rm ID_{l}\downarrow$ & $\rm CR_{r}\uparrow$ & $\rm CR_{l}\uparrow$ & $\rm IVU\downarrow$ & $\rm Phy\uparrow$ & $\rm SD\uparrow$ \\ 
\midrule
6&\textbf{2.68}&8.51&2.26&0.77&0.34&9.10&3.29&0.13&90.63&0.29\\
12&4.77&7.48&1.35&0.93&0.35&11.26&2.86&0.10&94.94&0.30\\
18&6.89 &\textbf{5.31}&1.24&\textbf{0.61}&\textbf{0.26}&11.38&\textbf{3.96}&\textbf{0.07}&98.25&\textbf{0.31}\\
24&9.26&5.48&\textbf{1.22}&\textbf{0.61}&0.31&\textbf{11.42}&3.54&0.08&\textbf{98.29}&\textbf{0.31}\\
\bottomrule
\end{tabular}
\label{tab:supmat_step}
\end{table}